\documentclass{article}

\PassOptionsToPackage{numbers, compress}{natbib}
\usepackage[final]{neurips_2019}

\usepackage[utf8]{inputenc} % allow utf-8 input
\usepackage[T1]{fontenc}    % use 8-bit T1 fonts
\usepackage{hyperref}       % hyperlinks
\usepackage{url}            % simple URL typesetting
\usepackage{booktabs}       % professional-quality tables
\usepackage{amsfonts}       % blackboard math symbols
\usepackage{nicefrac}       % compact symbols for 1/2, etc.
\usepackage{microtype}      % microtypography
\usepackage{graphicx}
\usepackage{caption}
\usepackage{subcaption}
\usepackage[ruled,noend, linesnumbered]{algorithm2e}

\usepackage{floatrow}
\newfloatcommand{capbtabbox}{table}[][\FBwidth]

\usepackage{blindtext}
\SetKwComment{Comment}{$\triangleright$\ }{}
\DeclareMathAlphabet\mathbfcal{OMS}{cmsy}{b}{n}

\title{Federated Uncertainty-Aware Learning for Distributed Hospital EHR Data}

\author{ %
{\centering  Sabri Boughorbel\textsuperscript{1,*}, Fethi Jarray\textsuperscript{2}, Neethu Venugopal\textsuperscript{1}, 
Shabir Moosa\textsuperscript{1}, Haithum Elhadi\textsuperscript{1}, } \\ \textbf{ Michel Makhlouf\textsuperscript{1} }  \\
  \textsuperscript{1}Sidra Medicine, Doha, Qatar, \textsuperscript{2}Higher Institute of Computer  Science, Medenine, Tunisia, \\
  \textsuperscript{*}\texttt{sboughorbel@sidra.org} \\
}

\begin{document}

\maketitle

\begin{abstract}
Recent works have shown that applying Machine Learning to Electronic Health Records (EHR) can strongly accelerate precision medicine. This requires developing models based on diverse EHR sources. Federated Learning (FL) has enabled predictive modeling using distributed training which lifted the need of sharing data and compromising privacy. Since models are distributed in FL, it is attractive to devise ensembles of Deep Neural Networks that also assess model uncertainty.  We propose a new FL model called Federated Uncertainty-Aware Learning Algorithm (FUALA) that improves on Federated Averaging (FedAvg) in the context of EHR. FUALA  embeds uncertainty information in two ways: It reduces the contribution of models with high uncertainty in the aggregated model. It also introduces model ensembling at prediction time by keeping the last layers of each hospital from the final round. In FUALA,  the Federator (central node) sends at each round the average model to all hospitals as well as a randomly assigned hospital model update to estimate its generalization on that hospital own data. Each hospital sends back its model update  as well a generalization estimation of the assigned model. At prediction time, the model outputs C predictions for each sample where C is the number of hospital models. The experimental analysis conducted on a cohort of 87K deliveries for the task of preterm-birth prediction showed that the proposed approach outperforms FedAvg when evaluated on out-of-distribution data. We illustrated how uncertainty could be measured using the proposed approach.

\end{abstract}

\section{Introduction}
Large volumes of EHR data are being generated during the daily clinical operation of hospitals. Developing Machine Learning models based on EHR has recently been very successful. In order to ensure good generalization of these models, a collective effort in required to include different data sources during model training. This will account for variability, data shift and potential biases across hospitals. EHR data sharing is very challenging due to the strict regulations on patient data privacy. Federated Learning (FL) paradigm has enabled the training of predictive models jointly from different EHR sources without the need for data sharing~\cite{FedertedConcept2017}. There are a few research challenges that face FL: statistical~\cite{FedertedRobust2019,li2019convergence}, communication bandwidth~\cite{McMahan2017,FedertedMT2017,konecny2016} and privacy~\cite{liu2018secure, cheng2019secureboost} challenges. 
% The statistical challenge concerns the fact that the samples do not follow the i.i.d. hypothesis (independent and identically distributed). Many theoretical foundations in machine learning are based on the i.i.d. Therefore FL needs new tools to  design theoretical sounds learning algorithms  \cite{FedertedRobust2019}. The communication challenge \cite{McMahan2017,FedertedMT2017,konecny2016}   deals with  reducing the communication volume  between the central server and the hospitals since there is large number of hospitals.  The privacy challenge ensures that the hospitals keep their data private during the model training process and that an attacker can not recover the local data by inspecting the model parameters.
%It  is governed by three parameters $(C,B,E)$ where  $C$ is the fraction of hospitals   updating the model. $B$ is the size of the mini-batch for each hospital and $E$ is the number of  epochs. 
The Federated Averaging (FedAvg) \cite{McMahan2017} is one the baseline algorithms in FL. FedAvg is  described as follows: At each communication round the server chooses a  C-fraction of hospitals to update the model. In turn, each selected hospital chooses a local mini-batch of size $B$ and runs $E$ epochs of Stochastic Gradient descent (SGD) and then sends back the updated model to the server to be averaged. In term of convergence,  FedAvg is practically equivalent to a central model  when i.i.d data is used. McMahn et al.~\cite{McMahan2017}  demonstrated that  FedAvg is still  robust for some examples of non-i.i.d. data.  However, Zhao et al.~\cite{zhao2018} showed that the accuracy of FedAvg is significantly reduced  when trained on highly skewed non-i.i.d. data even under convex optimization setting.\\
%As opposed to FedAvg, Guha et al. proposed single round communication algorithm where each hospital fits  a local model on its own local data and then the global server uses ensemble method to integrate these local models into a global model \cite{guha2019one}.\\
For EHR data,   hospital infrastructure, physicians and healthcare regulations influences directly the  choices of diagnosis, lab orders, treatments, etc. These factors are likely to vary across hospitals and hence invalidating the i.i.d assumption.  To tackle non-i.i.d. samples in FL,  Smith et al.~\cite{FedertedMT2017} proposed a multi-task learning framework. Zhao et al.~\cite{zhao2018} proposed a sharing strategy where only a  data subset is made available by each hospital. D. Liu et al.~\cite{liu2018fadl} proposed to train an aggregated FL model for EHR. The last layers which are trained specifically for each hospital data. This approach was not designed for out-of-distribution data and rather to be applied to each hospital own data.   Li et al.~\cite{li2019convergence} proved a convergence bound for FedAvg with different sampling and aggregation schemes under strongly convex and smooth functions. However, these assumptions are not valid in deep learning. 
%\cite{zhu2019multi} proposed a multi-objective evolutionary algorithm to minimize the global model test errors
%Li et al.~\cite{li2019convergence} improved FedAvg by a  partial device participation scheme where the central server aggregates only  the outputs of the first fraction of the hospitals.
In  the context of healthcare, solely relying on the prediction of ML models could  compromise patient safety. Deep learning models sometimes tend to have a high confidence in prediction as they can memorize training examples. Therefore it is highly desired to incorporate uncertainty notion in ML models for healthcare applications. Different approaches have been developed to measure uncertainty in deep learning models\cite{lakshminarayanan2017simple,dusenberry2019analyzing}. In this contribution, we focus on modeling uncertainty via model ensembling. Our base Neural Network is RETAIN which has two parallel RNN branches merged at a final logistic layer \cite{choi2016retain}.  Algorithm \ref{algo:fual} outputs the average model prediction  and the logistic layer for each hospital. To determine the uncertainty of a new patient, we collect the predictions of $C$  models. The final prediction is taken as the average of the $C$ scores. The prediction variance and other metrics can be used to assess the prediction uncertainty.
In this work, we assume that the communication bandwidth between hospitals is not a bottleneck.  We assume also that central server in FL is a trusted party.
\section{Clinical Application and Study Cohort}

In this paper, the clinical application is to predict preterm birth based on EHR data~\cite{verhaegen2012accuracy, kramer2013analyzing}.   Preterm birth is the leading cause of mortality in neonates and long-term disabilities. Between 9\% to 15\% of babies are born before 37 weeks of gestation~\cite{barros2015distribution}. The cost of their delivery and care exceed 26 billion dollars in the US \cite{behrman2006preterm}. The task is to predict the risk for a preterm-birth 3 months prior to delivery \cite{tran2016preterm, vovsha2016using, boughorbel2018alternating} using only EHR information. The dataset is extracted from Cerner Health Facts EHR database. The pregnancy episodes were identified  between 2014 and 2017. ICD-10 codes were selected to define the classes of preterm deliveries and full-term deliveries. The pregnancy timeline is anchored with respect to the delivery timestamp as there is no information available on the gestational age.  The data history used during training is restricted between 9 months before delivery and 3 months before delivery. The remaining data before and after this interval is discarded from the cohort. For this work, we extracted the top-50 hospitals having the most delivery episodes. The clinical information included in the data are: diagnosis information (ICD-9 and ICD-10), medication orders (NDC), lab orders (LOIN-C), procedures (ICD-10, ICD-9), surgery, micro-biology information. In total, the cohort contained  87,574 deliveries with their history between 9 and 3 months prior to delivery. Each sample is represented as a sequence of clinical events \cite{choi2016doctor} associated with a label indicating preterm or full-term delivery .  Table \ref{tab:cohort} gives a summary statistics about the study cohort.

\begin{table}[h]
    \centering
    \caption{The cohort data is extracted from Cerner EHR database between 2014 and 2017. The prevalence of preterm in the cohort is 7.6 \%.}
    \begin{tabular}{lcc} \toprule
Information & full-term & pre-term \\ \midrule
\# deliveries & 80,900 & 6,674 \\ 
Age         & 28.4$\pm$5.7 &29.2$\pm$5.9 \\ 
\# encounters & 6.4$\pm$3.9 & 6.8$\pm$4.7 \\ 
 \bottomrule 
    \end{tabular}
    \label{tab:cohort}
\end{table}
\section{Method}
The EHR dataset contains structured data. Each subject data  is represented as a sequence of discrete clinical codes. We used RETAIN model for the task of predicting preterm-birth 3 months before delivery we trained Recurrent Neural Networks models on each hospital data in a federated learning setup. \cite{choi2016retain}. RETAIN has an embedding layer and two RNN branches with attention mechanisms. This enables the interpretability of the model by quantifying the importance of features as well as the subject visits. It has been successfully used for EHR data for clinical applications.

%Then we present the baseline federated learning model and give details on the proposed method. 
%\subsection{Deep Learning Model}
\subsection{Model Sampling}
We propose an improvement to FedAvg to incorporate uncertainty modeling.  We introduce a criterion for hospital model sampling which diminishes the contribution of models with low prediction performance.  The latter is measured as a generalization estimation on out-of-distribution data. As proposed in \cite{li2019convergence},  sampling can reduce the influence of non-i.i.d. during federated model training. The weights in the sampling are usually set to be the data size of each hospital. We propose to use hospital model generalization as weights for  sampling. Hence we favor models with good generalization to be selected for model averaging. 
%Uncertainty is also introduced in the model by. 
In addition, the model updates from hospitals can be used to introduce a model ensembling and derive uncertainty measure at prediction time.
\subsection{FUALA: Federated Uncertainty-Aware Learning Algorithm}
Algorithm \ref{algo:fual} provides a pseudo-code for the proposed method. We define  $P_k^t \in \{1, \cdots, K\}$ as a permutation vector on the set of $K$ hospitals at round $t$. The permutation is only known to the Federator (central node) and it is randomly changed at each round. The model associated with hospital $P_k^t$ will be evaluated on the data from hospital $\mathcal{H}_k$. In this way, the model updates are sent to the data location and hence preserves data privacy in the algorithm.

\begin{algorithm}[H]
\label{algo:fual}
\SetKwInOut{Input}{Input}
\SetKwInOut{Output}{Output}
%\underline{function Euclid} $(a,b)$\;

\Input{ $\mathbfcal{F}$: Federator (central server), \, $\mathbfcal{H}$: Hospital (hospital server), \, $\mathbf{K}$:\# of hospitals, \, $\mathbf{C} \leq K$: \# of selected hospitals, \, $\mathbf{E}$:, \# of local epochs, \, $\mathbf{T}$: \# rounds. }
   
$\mathcal{F}$ initializes sampling weights as $h_k = \frac{1}{K}$ \\
$\mathcal{F}$ initializes the model  $w^{0}$ sends it to $\mathcal{H}_{1,\cdots, K}$ ($w^{0}_k=w^{0}$)  \\
%$\mathcal{F}$ sets randomly the permutation vector $P^{0}$  \\
%$\mathcal{H}_k$ initializes its local model  $w^{0}_k=w^{0}$ \\
 \For{ each round t=0,1, \ldots,T}{
    $\mathcal{F}$ sets randomly the permutation vector $P^{t}$   \Comment*[r] {for $\mathcal{H}_k$  to evaluate the model of   $\mathcal{H}_{P_k^{t}}$}
   $\mathcal{F}$ selects a C-fraction  $S_{t}$ of $\mathcal{H}_{1,\cdots, K}$ \Comment*[r]{randomly  selected, weighted by $h_{k}$ for  $\mathcal{H}_k$} 
   $\mathcal{F}$ sends the model $w^{t}$ to $\mathcal{H}_{1,\cdots, K}$ \\
    $\mathcal{F}$ sends  $w_{P_k^t}^{t}$ to $\mathcal{H}_k, k=1, \ldots,K$ \\
       %$a_{P_k^t},k=1,2, \ldots,K $  \\
   \For{ each $\mathcal{H}_k$, k=1,2, \ldots,K}{
    $\mathcal{H}_k$ updates $w^{t}$ for $E$ epochs to obtain $w_{k}^{t+1}$ and sends it back to $\mathcal{F}$ \;
    $\mathcal{H}_k$ computes the  generalization  $a_{P_k^t}^t$ of model $w_{P_k^t}^{t}$ \;
     }
   $\mathcal{F}$   updates the sampling weights as  $h_{k}^{t+1}=a_{k}^t  + h_{k}^t, k=1,\ldots, K$ \; 
   $\mathcal{F}$ aggregates $w_k^t$ as $w^{t+1}= \displaystyle \frac{1}{C}\sum_{k\in S_{t}}w_{k}^{t+1}$;
  }
\Output{ Average model $w^T$}
 \caption{FUALA: Federated Uncertainty-Aware Learning Algorithm}
\end{algorithm}
%\end{minipage}

% how to treat the case when a hospital is selected twice. It may happen because we use repetition

   At each round $t$, the Federator sends the average model $w^t$ to all hospitals. In addition, it assigns hospital $\mathcal{H}_k$ the model ($w_{P_k^t}^{t}$) for evaluation.  The task of hospital $\mathcal{H}_k$ is to update the average model $w^t$ for $E$ epochs on its local data and to estimate the performance of model $w_{P_k^t}^{t}$ also on its own data. The data of hospital $\mathcal{H}_k$ can be viewed as a test set for hospital $\mathcal{H}_{P_k^t}$ . We chose $a_k$ = (AU-ROC+PR-AUC)/2 as the generalization metric. After a few rounds, the generalization estimation improves as different hospital data is assigned for each model evaluation.
   %By assigning a single model at each round we keep limit the communication bandwidth as well as risk of privacy leakage.  The latter can be further improved by clustering hospital models together and applying the model average. The generalization scores will be assigned for the cluster instead. Although this step increases the communication cost from the server to hospitals, we consider the hospital infrastructure allows this assumption. Another more costly option is to send more than models for the generalization estimation. Each hospital  will have the task to update the average and evaluate the generalization of the received models on its own data. We considered  as a generalization metric. The central server obtains the updated models and estimated generalization weights. These weights are then used in hospital sampling. 
%\vspace*{-0.35cm}
\section{Experiments and Uncertainty Evaluation}
\begin{figure}
\begin{floatrow}
\ffigbox{%
  %\rule{3cm}{3cm}%
  \includegraphics[width=0.8\linewidth]{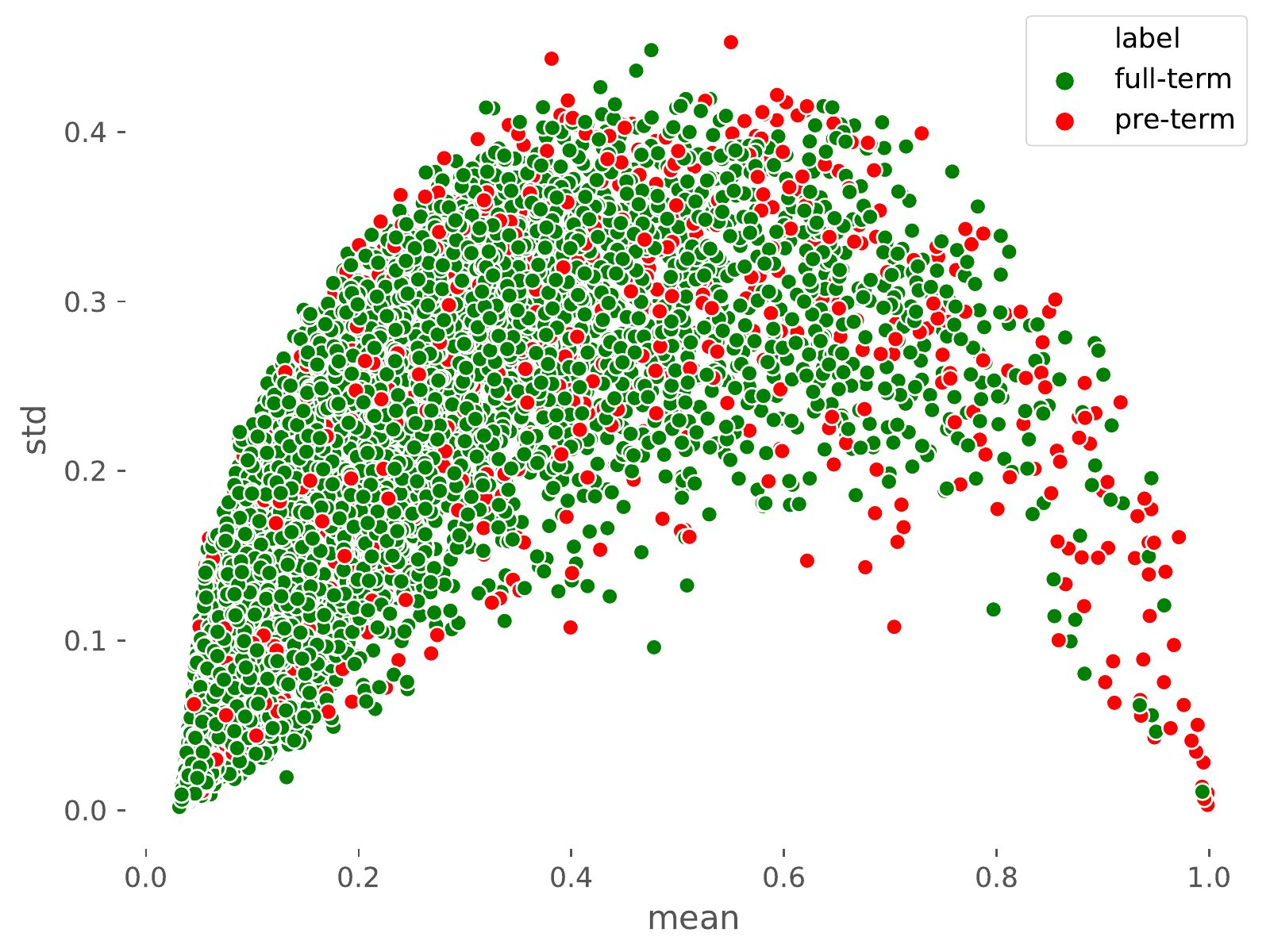}
}{%
  \caption{std vs. mean (prediction) of the models.}%
  \label{fig:std}
}
\capbtabbox{%
\begin{tabular}{lcc} \toprule
Model & AU-ROC & AU-PR \\ \midrule
Central Model  & 72.5$\pm$0.8 &21.8$\pm$1.0 \\ 
FUALA & 67.8$\pm$2.7 & 19.6$\pm$4.3\\   
Weight. FedAvg & 63.5$\pm$2.7  &14.7$\pm$3.0 \\ 
FedAvg    &62.2$\pm$2.2 & 18.6$\pm$2.9\\  \bottomrule
\end{tabular}
}{%
  \caption{Performance comparison.}%
  \label{tab:performance}
}
\end{floatrow}
\end{figure}

\begin{figure}
\centering
\begin{subfigure}{.5\textwidth}
  \centering
  \includegraphics[width=0.8\linewidth]{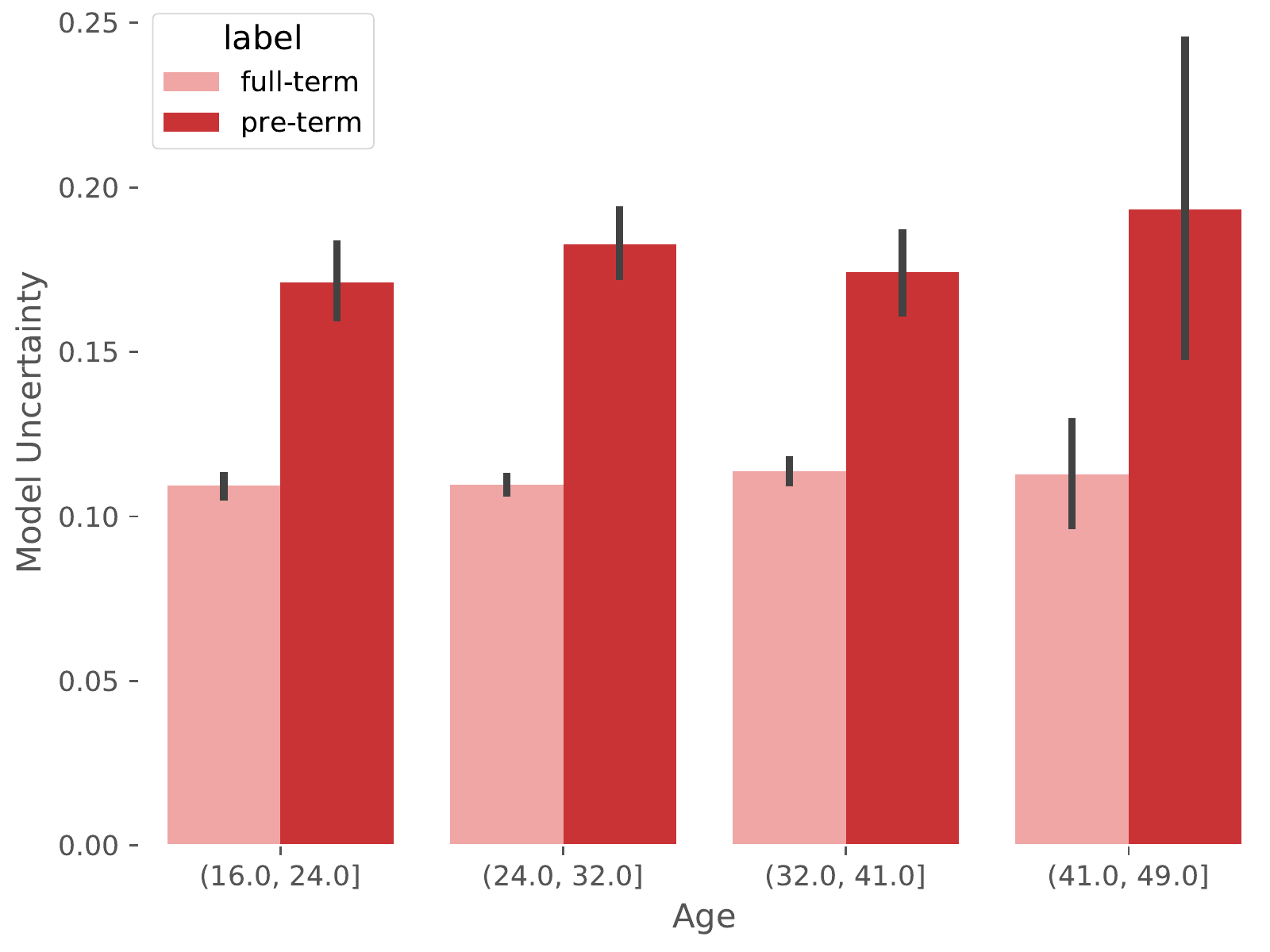}
  \caption{Model uncertainty as a function of the mother's age.}
  \label{fig:age}
\end{subfigure}%
\begin{subfigure}{.5\textwidth}
  \centering
  \includegraphics[width=0.8\linewidth]{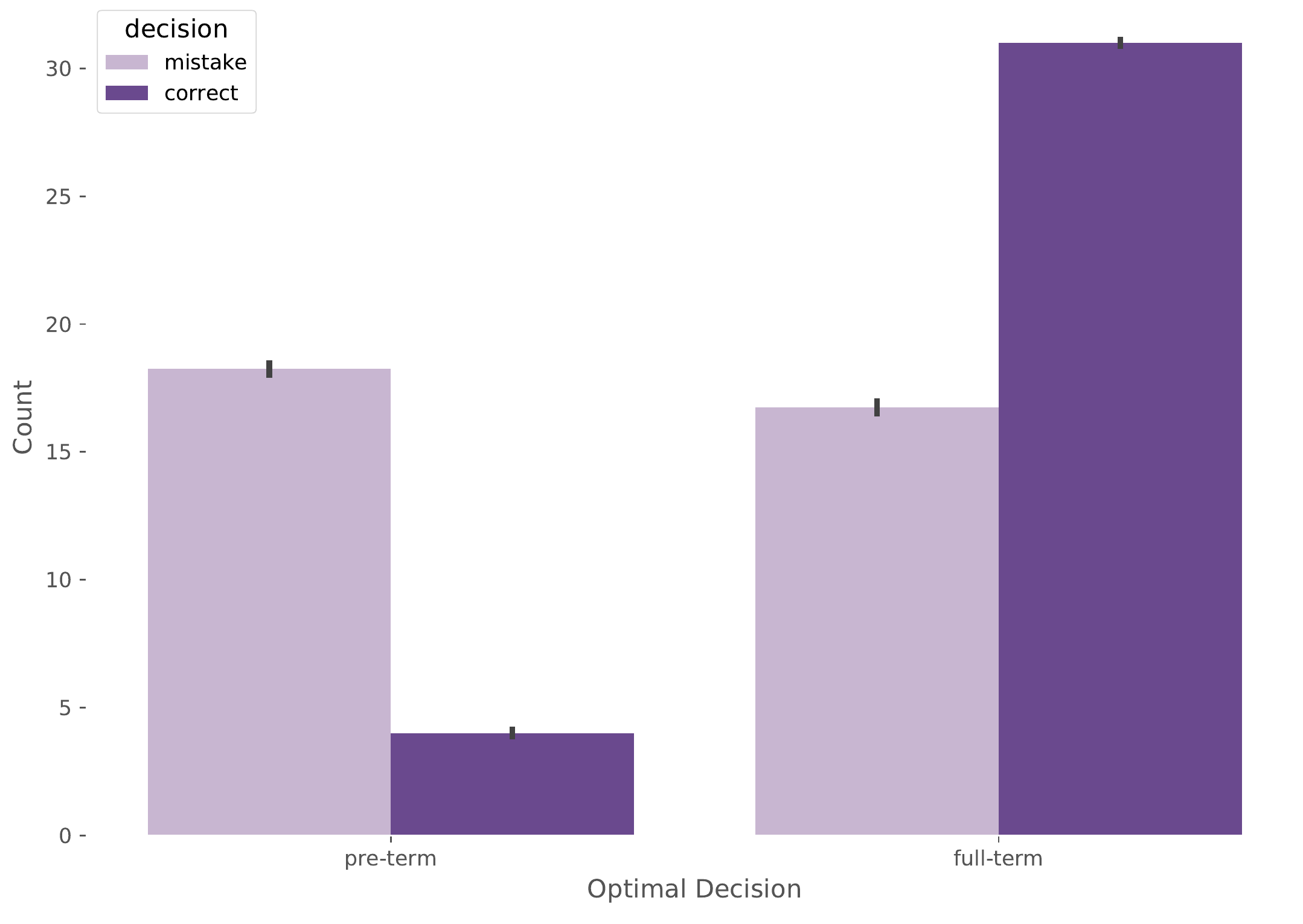}
  \caption{Count of model decisions for the two classes. \label{fig:optimal}}
\end{subfigure}
\caption{Performance results and uncertainty analysis based on the test set.}
\label{fig:results}
\end{figure}

The EHR data from $K$=42 hospitals are used to train the FL models and 8 for testing.  This is repeated 10 times by randomly selecting different training and test sets. The mean and variance performances are summarised in Table~\ref{tab:performance}. The central model (no FL) had the best performance where all hospital data are merged and uniformly sampled. This result is expected since the data distribution of the training and test sets are the same. For the FL approaches, we set, 20 rounds and 5 epochs per update. The fraction  $C$=35 is chosen. Instead of using hospital data size as averaging weight in FedAvg, we chose the normalized $a_k$ (model generalizations) of each hospital as weights. This model is denoted by weighted-FedAvg in Table~\ref{tab:performance}. It has slightly better performance than FedAvg. The proposed FUALA outperformed the two previously mentioned FL methods (FedAvg and weighted FedAvg).
% \subsection{Uncertainty  Evaluation}
We conducted a few experiments to assess the uncertainty in FUALA. The model outputs 35 predictions from the final ensemble model. The std and mean of predictions are depicted in Figure~\ref{fig:std}. Predictions near zero and one have low variance and hence have a high confidence. Figure \ref{fig:age} shows the variation of model uncertainty (prediction std) as a function of age. Pre-term prediction has a higher uncertainty than full-term for all age groups.  The age group > 41 has  a high uncertainty for preterm prediction. Figure \ref{fig:optimal} illustrates how the disagreement in the ensemble models for pre-term and full-term classes and for correct and false classification. Among the 35 model decisions, about 17 models voted for one of the classes in case of mistakes. Thus the model shows high uncertainty when making mistakes. In our experiments, the model prediction was based on the average probability prediction from the ensemble. FUALA could be improved by adding a voting mechanism for the final prediction. The results in Figure \ref{fig:optimal} could be also used to define a rejection procedure when the disagreement between model decisions is high.
% \label{results}
% \begin{table}
% \centering
% \begin{tabular}{lcc} \toprule
% Model & AUC-ROC & AUC-PR \\ \midrule
% Central Model  & 72.5$\pm$0.8 &21.8$\pm$1.0 \\ 
% FUALA & 67.8$\pm$2.7 & 19.6$\pm$4.3\\   
% Weighted FedAvg & 63.5$\pm$2.7  &14.7$\pm$3.0 \\ 
% FedAvg    &62.2$\pm$2.2 & 18.6$\pm$2.9\\  \bottomrule
% \end{tabular}
% \caption{Performance comparison of the experimental results.}
% \label{tab:results}
% \end{table}

%\vspace*{-0.35cm}

\section{Conclusion}
In this paper we proposed FUALA a federated uncertainty-aware learning algorithm for the prediction of preterm birth from distributed EHR. Our contribution is twofold. First, we used the generalisation performance of each hospital model as sampling weight. Second, we introduced model ensembling by keeping the last layer of each model to measure the uncertainty of the final model.  We showed that FUALA have good performance compared with other FL methods. We illustrated how  model uncertainty can be measured. In a future work, we plan to apply the model to a larger dataset and design a rejection criterion.

\bibliography{ml4h_bibliography.bib}

\end{document}